\documentclass[conference]{IEEEtran} 

\usepackage{makecell}
\usepackage{multirow}
\usepackage{float}
\usepackage[linesnumbered,ruled,vlined]{algorithm2e}
\SetKwInput{KwInput}{Input}   
\SetKwInput{KwOutput}{Output} 

\usepackage{hyperref}
\usepackage{amsmath,amsfonts,amssymb}
\usepackage{graphicx}
\usepackage[numbers]{natbib}

\usepackage[caption=false,font=footnotesize]{subfig}

\hypersetup{
  colorlinks   = true, 
  urlcolor     = black, 
  linkcolor    = black, 
  citecolor   = black 
}

\begin{document}
%
%
%
%
\title{EsmamDS: A more diverse exceptional survival model mining approach}
\author{\IEEEauthorblockN{Juliana Barcellos Mattos}
\IEEEauthorblockA{\emph{Centro de Informática (CIn)} \\
\emph{Universidade Federal de Pernambuco}\\
Recife, Brazil \\
julianabarcellosmattos@gmail.com} 
\and
\IEEEauthorblockN{Paulo S. G. de Mattos Neto}
\IEEEauthorblockA{\emph{Centro de Informática (CIn)} \\
\emph{Universidade Federal de Pernambuco}\\
Recife, Brazil \\
psgmn@cin.ufpe.br} 
\and
\IEEEauthorblockN{Renato Vimieiro}
\IEEEauthorblockA{\emph{Depto. de Ciência da Computação (DCC)} \\
\emph{Universidade Federal de Minas Gerais}\\
Belo Horizonte, Brazil \\
rvimieiro@dcc.ufmg.br} 
}

\maketitle
%
%
%
\begin{abstract}
A variety of works in the literature strive to uncover the factors associated with survival behaviour.
However, the computational tools to provide such information are global models designed to predict 
\emph{if} or \emph{when} a (survival) event will occur. When approaching the problem of explaining differences in survival behaviour, 
those approaches rely on (assumptions of) predictive features followed by risk stratification.
In other words, they lack the ability to discover new information on factors related to survival.
In contrast, we approach such a problem from the perspective of descriptive supervised pattern mining to discover local patterns associated with different survival behaviours. Hence, we introduce the EsmamDS algorithm: an Exceptional Model Mining framework to provide straightforward characterisations of subgroups presenting unusual survival models -- given by the Kaplan-Meier estimates.
This work builds on the Esmam algorithm to address the problem of pattern redundancy and provide a more informative and diverse characterisation of survival behaviour.
\end{abstract}

%
%
%
\section{Introduction}
\label{sec:intro}

We investigate in this work how to find exceptional subgroups in \emph{time-to-event} or \emph{survival} data.
Survival data arises when the outcome of interest in an experiment is not only \emph{if} an event has occurred, but also \emph{when} it occurred. There exists a broad range of problems that fall into this class.
For instance, one might be interested in knowing not only \emph{if} a patient relapsed (event) but also \emph{when} it occurred. 
The same outcome would be expected by a company evaluating customer churn.
The area that deals with problems like these is known as \emph{Survival Analysis} (SA).

One of the main goals of survival analysis is to create a model that is able to answer the \emph{if} and \emph{when} questions
usually posed in these studies. In this sense, one might think that a natural solution would be to state the problem as a simple regression task. 
This, however, is not a proper solution since traditional methods are unable to consider censored cases, that is instances for which the occurrence of the event is unknown due to time restriction or losing track during the study~\citep{SA:survey_MachineLearning-survival}.
For this reason, many studies focus on adapting traditional
machine learning methods for survival data analysis as pointed out by \citet{SA:survey_MachineLearning-survival}.
In other words, the main goal of machine learning methods for survival analysis has been on creating a predictive model for the data.

Although predictive models are important in many contexts, they fail at identifying features that might affect survival for a subgroup of
instances. That is, as they attempt to only answer the \emph{if} and \emph{when} questions, they fail to answer questions such
as `\emph{are there groups of instances with unusual survival?}' or `\emph{what features are associated with this exceptional behaviour?}'.
This situation is not a problem per se, however, the lack of methods that answer these more descriptive questions highlights the gap
in the literature. And this motivates our work.

We see the problem from the descriptive perspective. We are interested in investigating whether the last two questions could also be
answered by data mining or machine learning methods. Traditionally, similar questions are answered by stratifying data regarding a
variable of interest. For instance, to verify whether a new treatment affects prognosis, patients may be split into two groups, those
who took the new drug, and those who took a placebo. Then, each group would be individually modeled and their differences analysed.
Notice, however, that such an approach assumes that the variables of interest are known a priori. This also is not always true, since
in some cases researchers may in fact want to identify these variables. \citet{MED:article_Basal-breast-cancer}, for instance, analysed
genes associated with a subtype of breast cancer that affected patient's survival.
They proposed a naive univariate approach to evaluate the effect of each individual gene on survival. 
Alternately, semi-parametric methods, such as Cox Proportional-Hazard models, may be used to assess the interaction between covariates~\citep{SA:survey_MachineLearning-survival,moncada2021}.
Both approaches have drawbacks; while the former has high computational costs and loss of interesting information as it neglects possible interactions between variables, the latter still focuses on a predictive model, thus, local patterns might be missed.

Alternative approaches were proposed by \citet{my:bracis} and \citet{SA:article_SD}. Both teams tackled the issues raised before 
by stating the problem as a \emph{supervised descriptive pattern mining} task. \citet{SA:article_SD}
proposed to tackle it with a \emph{subgroup discovery} (SD) algorithm, in which the average survival time was considered the target, 
while \citet{my:bracis} proposed an \emph{exceptional model mining} (EMM) algorithm to find itemset induced subgroups on 
which a non-parametric survival model was adjusted and compared against a baseline model of the complement. 

Although these methods solved in some extent the problems raised before, they still presented their on issues. Approaching the problem
as a standard SD task may lead to the loss of interesting patterns due to outliers in the subgroups. The method proposed by
\citet{my:bracis}, on the other hand, is very immature. First, in our preliminary studies, we observed that the patterns returned by
the algorithm were extremely redundant. Second, they only considered the comparison of a subgroup against its complement,
however, as discussed by \citet{EMM:article_EMM-Complex-Target-Concepts}, quality metrics distinguishing subgroups from their complements,
or subgroups and the whole data may yield different results. 

We address these issues in this work and propose a new EMM method for finding diverse subgroups with unusual survival behaviour.
We follow the discussion presented by \citet{SD:algorithm_DiverseSD} and tackle both coverage and description redundancy,
and also addressed a new type of redundancy: \emph{model redundancy}. Our method also allows the user to choose between
complement or whole data (called population in this work) comparisons. We also enhance the description language 
to increase generality and expressivity of the final set of patterns.
To evaluate the performance of our proposal, we conduct experiments with a set of real-world data. We compare our method to
the state-of-the-art algorithms in terms of the exceptionality, complexity, generality and redundancy of the final set of patterns.

The remainder of this work is organised as follows.
In \autoref{sec:preliminaries}, we introduce the main concepts of Survival Analysis and Exceptional Model Mining.
In \autoref{sec:algorithm}, we present the EsmamDS algorithm, followed by \autoref{sec:experiments}, which presents the empirical evaluation of our proposal.
Finally, in \autoref{sec:conclusions}, we draw some conclusions and present directions to extend this proposal.
%
%
%
\section{Preliminary concepts}
\label{sec:preliminaries}

In this section we briefly introduce some concepts necessary to understand our approach. We start with an introduction of
survival analysis and then present definitions related to subgroup discovery and exceptional model mining.

We call an \emph{event} any designated experience of interest that may happen to an individual under study.
A device failure, the period a user stays on a website, and death are all examples of events.
Often in survival analysis there is interest in predicting both \emph{if} and \emph{when} an event occurred.
One difficulty in real-world situations, however, is that not all the individuals will experience the event during the course of study,
or may even withdraw from it before its end. So data will be available only for a subset of the individuals in the study.
This phenomenon is called \emph{censoring}.

Let $A_1, A_2, \dots, A_n$ be a set of $n$ categorical attributes each with domain $dom(A_i)$,  $T$ the time-to-event (survival time) numeric 
attribute (the set of survival times of all individuals in the experiment), and $\delta$ the event indicator (survival status). 
An individual  $o$ is described by a set of $n$ values from the domain of each attribute,
has a survival status $\delta_o$ and with time $T_o$. Each individual is therefore a tuple $o = (V,T_o,\delta_o)$, where $V \in dom(A_1)\times 
dom(A_2)\times \dots \times dom(A_n)$; the value of an attribute $A_i$ is also denoted by $A_i(o)$. 
We define a (survival) data set $\Omega$ as a collection of individuals.
At last, we call the set of all attribute values the \emph{items} of the data set and denote it by $\mathbb{I} = \bigcup dom(A_i)$.

A survival function $S(t)$ is a representation of the probability of an individual surviving up to a time $t$, i.e., $S(t) = P(T > t)$.
As we previously stated, the survival function may be modeled by different methods. One of the simplest methods to estimate
such a function is by the non-parametric Kaplan-Meier (KM) estimator.
The KM model estimates the survival function by calculating the cumulative survival probability $\hat{S}(t)$ from the observed 
survival times $T$ and censoring information $\delta$.
The estimated probability $\hat S(t_j)$ of surviving past a time $t_j \in T$ is given by \autoref{eq:KM-simplify-formula}:

\begin{equation}
\label{eq:KM-simplify-formula}
\hat{S}(t_j) = \hat{S}(t_{j-1}) \left( 1 - \frac{d_j}{r_j} \right)
\end{equation}
where $\hat S(t_{j-1})$ is the survival probability at the time $t_{j-1}$, $r_j$ is the number of individuals  at risk at time $t_j$, 
i.e. $r_j = |\{o \in \Omega ~|~ T_o > t_j\}|$, and $d_j$ is the number of individuals that experienced the event at time $t_j$,
that is $|\{o \in \Omega ~|~ T_o > t_j \land \delta_o = 1\}|$.
The plot of the KM survival probabilities $\hat{S}(t)$ against time is called survival curve and provides a visual assessment of the survival response over time.
Two curves can be compared to verify whether they are statistically equivalent or not.
The \emph{log-rank} tests the null hypothesis that there is no difference between the two populations in the probability of an event occurring
at any time point~\citep{bland2004}.

We call a \emph{description} a conjunction of conditions on the attributes of the data set. Each attribute may be constrained at most once.
Thus, a description might have at most $n$ conditions. 
In this work, we define a condition as a restriction on the values that the
attribute might have. 
Thus, it can be seen as an indicator function $D: \Omega \to \{0,1\}$ defining whether an individual in the data set satisfies or not the constraints in the descriptions.
For instance, $D(o) = A_i(o) \in \{v_{i_1}, v_{i_2}, \dots, v_{i_p} \} \land A_j(o) \in \{v_{j_1}, v_{j_2}, \dots, v_{j_q} \} \land
A_k(o) \in \{v_{k_1}, v_{k_2}, \dots, v_{k_r} \}$ is a description of size 3 (it has 3 conditions), where the values for the attribute $i$ are restricted to those $p$ values in the first set, the values for $j$ are restricted to those $q$ values in the second, and the values for attribute $k$ are restricted to those $r$ values in the third set. 
In this case, $D$ returns 1 for individual $o \in \Omega$ if its attribute values are within the restricted values for each condition, and 0 otherwise.

A description $D$ \emph{covers} an individual $o \in \Omega$ if $D(o) = 1$.
We call a \emph{subgroup} the set of individuals covered by a description. Thus, the subgroup associated with a description $D$ is $G_D = \{o \in \Omega ~|~ D(o) = 1\}$ and the \emph{coverage} of $D$ is $|G_D|$.
On the other hand, the set of individuals not covered by $D$ is call the \emph{complement} of its subgroup (denoted by $\overline{G_D}$).
We call the \emph{items} of a subgroup the set of attribute values of its individuals. More formally, the items of a subgroup is a function
$\gamma: \mathcal{P}(\Omega) \to \mathbb{I}$, $\gamma(G_D) = \bigcup \{A_i(o) ~|~ o \in G_D\}$.
Note that may include even attribute values not listed in the conditions of the description that originated the subgroup. 
We use $\mathcal{I}(G_D)$ to represent the set of items strictly listed in the subgroup's description, i.e. the union of all constrained $dom(A_i)$ in $D$.

Let $\mathcal{D}$ be the set of all possible descriptions in a data set. We assess the interestingness (exceptionality) of
a subgroup $G$ induced by a description $D$ with a \emph{quality measure} $\phi: \mathcal{D} \rightarrow \mathbb{R}$.
This function quantifies how much the distribution of a target variable in the subgroup differs from the distribution in a \emph{baseline} group.
There are two possibilities in this case: (i) to compare the distribution of the target variable in the subgroup to its \emph{complement}; 
or (ii) to compare to the \emph{population}, i.e. the distribution for the whole data set.
There is no overall best choice.
However, it is crucial to understand that this choice essentially changes the nature of the task at hand and may lead to different outcomes.
Comparing a subgroup to the population implies searching for deviations from the norm.
On the other hand, comparing to its complement implies searching for two subsets presenting contrasting behaviour.
We implemented our approach to perform both baseline comparisons, which we evaluate separately in our experiments.
From now on, we refer to a general \emph{baseline} $\mathcal{B}$ as a predefined baseline target to quantify the exceptionality of a subgroup: \emph{population} or \emph{complement}.

We define exceptionality in this work by means of the log-rank statistic. For this we test the hypothesis that the KM curves adjusted
for both the subgroup and the baseline are statistically equivalent. Then, we take $1-$ \emph{p-value} of test as the quality of the subgroup.
The stronger the evidence that the null hypothesis should be rejected, the more exceptional the subgroup is.
This quality measure is presented in \autoref{eq:quality_emm}, where $p_{KM}$ is the \emph{p-value} of the log-rank test.
\begin{equation}
\label{eq:quality_emm}
\phi(G,\mathcal{B}) = 1 - p_{KM}
\end{equation}

We now proceed to describing how exceptional subgroups are found, and present our method.
%
%
%
\section{EsmamDS: Exceptional Survival Model Ant-Miner - Diverse Search}
\label{sec:algorithm}

The Exceptional Survival Model Ant-Miner Diverse Search (EsmamDS) is an EMM framework that extends the work presented 
by \citet{my:bracis} to provide a set of more diverse subgroups. Our method tackles redundancy in three different dimensions: description; 
coverage; and survival model.
The EsmamDS algorithm employs the Kaplan-Meier Estimates as the target model and the quality measure $\phi$ defined in \autoref{eq:quality_emm}.
Instead of defining a baseline model to compare subgroups with, EsmamDS provides this choice as a user-defined parameter.
The framework's pseudocode is provided in \autoref{alg:esmam_framework}.
\begin{algorithm}[t]
    \scriptsize
    \caption{EsmamDS Framework}
    \label{alg:esmam_framework}
    \DontPrintSemicolon
      
    \KwInput{$\mathcal{B}$ -- baseline for subgroup comparison, $\alpha$ -- level of significance, $maxStag$ -- maximum stagnation of the algorithm, $\wp^S = \{nAnts, nConverg$, $minCov\}$ -- hyperparameters of the subgroup search, $\wp^H =\{L,W\}$ -- hyperparameters of the heuristic function}
    \KwOutput{$\mathbb{G}$ -- set of exceptional subgroups}
    \KwData{$\Omega$ -- survival data set }
  
    $\mathbb{G} \leftarrow \varnothing$; $G\leftarrow \varnothing$\\
    $\mathbb{U} \leftarrow \Omega, \Delta U \leftarrow 0$ \\
    $stag \leftarrow 0$
    
    \While{$\mathbb{U} \neq \varnothing$ \textbf{and} $stag \leq maxStag$}
    {
   		\texttt{searchInitialisation}$(\mathcal{I}(G), \mathbb{G}, \mathbb{I}, \mathbb{U}, \wp^H)$\\
   		$G \leftarrow$ \texttt{subgroupSearch}$(\mathcal{B}, \wp^S)$\\
   		$\mathbb{G} \leftarrow$ \texttt{subgroupSetUpdating}$(G, \mathbb{G}, \alpha)$\\
   		
   		$\Delta U \leftarrow |\mathbb{U}| - |\bigcup_{G^k \in \mathbb{G}} G^k|$\\
   		$\mathbb{U} \leftarrow \bigcup G^k$\\
   		
       	\If{$\Delta U = 0$}{
       	    $stag \leftarrow stag + 1$
       	}
       	\textbf{else:} $stag \leftarrow 0$\\
   		
    }
    \textbf{return:} $\mathbb{G}$
\end{algorithm}

The algorithm is initialised with an empty set of discovered subgroups $\mathbb{G}$, an empty subgroup $G$ (note that $G=\varnothing \rightarrow \mathcal{I}(G)=\varnothing$), and an initial set of uncovered observations $\mathbb{U}$ containing all individuals in $\Omega$.
Then, following a covering-based approach (lines 4-12), it iteratively searches for subgroups (considering a baseline $\mathcal{B}$) until all observations in $\Omega$ are covered by $\mathbb{G}$ at least once, or until the algorithm achieves a maximum stagnation threshold ($maxStag$) -- i.e. a number of consecutive iterations with no change in $\mathbb{U}$.
The process of discovering subgroups uses the Ant-Colony Optimisation (ACO) metaheuristic as search strategy to construct descriptions from the search space defined by $\mathbb{I}$.

In ACO metaheuristic, a colony of artificial ants constitute an iterative procedure that stochastically constructs solutions (descriptions) given a probability distribution associated to solution components -- here, the items $I_{ij} \in \mathbb{I}$, for attribute $A_i$ and value $v_{j} \in dom(A_i)$.
Each individual ant builds a complete description through a probabilistic choice of solution components, and exchange information on the solution's quality through a pheromone trail -- an indirect form of communication that allows the algorithm’s search experience to bias the solution construction of future ants.
Such probabilistic choice is influenced by the amount of pheromone associated with available items and by an heuristic information about the problem, which helps to guide the search towards the most promising solutions.

Finally, the EsmamDS framework consists of three major steps:
(i) the initialisation of the probabilistic elements of the ACO search (line 5);
(ii) the subgroup search, which returns the best subgroup $G$ discovered by a complete colony of ants (line 6);
and (iii) the update of the final subgroup set $\mathbb{G}$ considering the discovered subgroup $G$ (line 7).
We now discuss each of these steps in detail. 

\subsection{Search Initialisation}
The \texttt{searchInitialisation} function is responsible for the initialisation of the pheromone values $\tau$ and heuristic values $\eta$ associated with each item $I_{ij} \in \mathbb{I}$.

At the beginning of each colony process, no pheromone has been  deposited in the trails yet, and all solution components receive the same pheromone values.
One should notice, however, that pheromone values are adjusted according with the colony's iteration: each ant in the colony updates the values to be used by the next ant.
The updating method will be presented in the next section along with the colony's searching process. Since the values are updated with iterations, we use superscripts to refer to the pheromone values in a specific colony iteration, for instance, $\tau^t$ denotes the pheromone value in iteration $t$.
Hence, here we define the initial configuration of the pheromone values as $\tau^0(I_{ij})=|\mathbb{I}|^{-1}$ for each $I_{ij} \in \mathbb{I}$.
Once the probabilistic choice of solution components is defined by both pheromone and heuristic values, and given that all pheromone values are equal at the beginning of each colony, we have that the heuristic associated to the items define the initial probability distribution of the search space.

In contrast to the static heuristic function proposed by \citet{my:bracis}, we propose a dynamic function $\eta: \mathbb{I} \rightarrow [0,1]$, $\eta(I_{ij}) = \eta_{H}(I_{ij}) \cdot \eta_{L}(I_{ij}) \cdot \eta_{W}(I_{ij})$.
In addition to use information theory to provide a problem-dependent quantification of the relevance associated with the items in the search space ($\eta_H$), we use both the descriptions and coverages of the discovered subgroups to improve search exploration ($\eta_L, \eta_W$).

The entropy-based component $\eta_{H}$ provides a quantification of the discriminative power of the items regarding survivability, and it is defined in \autoref{eq:alg_heuristic_entropy} as
\begin{equation}
\label{eq:alg_heuristic_entropy}
\eta_H(I_{ij}) = \frac{\log_2 k - H(W|I_{ij})}{\sum\limits_{I_{ij} \in \mathbb{I}} \log_2 k - H(W|I_{ij})}
\end{equation}
where $H(W|I_{ij}) = -  \sum_{w=1}^{k} P(w|I_{ij}) \cdot \log_2 P(w|I_{ij})$ is the Shannon's entropy.
We considered an initial partition of the individuals in $\mathbb{U}$ as those with survival time at least as long as the data set average survival time, and those with shorter survival time.
Note that in each iteration of the algorithm, the $\eta_H$ heuristic is computed over the set of observations that remain uncovered by the subgroups in $\mathbb{G}$.
The more uniformly distributed is an item across those two survival groups (i.e. if it does not discriminate between the considered survival partition), the smaller is its heuristic quantification.
By contrast, $\eta_H(I_{ij})$ assumes maximum value when the item is fully associated with a  single survival group.

The description attenuation component $\eta_{L}$ uses all the items in previously discovered subgroups to guide the search towards unvisited (or more rarely visited) items in the search space.
The rationale is that more discriminative items may bias the search, and, thus, we should penalise them somehow to promote diversity.
Such a penalisation is based on the logistic function as presented in \autoref{eq:alg_heuristic_descrip},
\begin{equation}
\label{eq:alg_heuristic_descrip}
\eta_{L}(I_{ij}) = 1 - \frac{1}{1+e^{-(c(I_{ij}) - L)}}
\end{equation}
where $c(I_{ij})$ is the number of times $I_{ij}$ was represented by previously discovered subgroups.
The parameter $L$ adjusts the penalisation of an item regarding its usage, defining the value of $c(I_{ij})$ for which the heuristic quantification $\eta(I_{ij})$ decreases by half.
Therefore, we have that the more an item appears in the descriptions discovered by the search, the smaller becomes its (\emph{a priori}) probability of being explored by future ant colonies.

Lastly, the weighted covering component $\eta_{W}$ uses the subgroups in $\mathbb{G}$ to guide the search towards items describing observations less represented in the final set.
For that, we make use of a score based on multiplicative weighted covering proposed by \citet{SD:algorithm_DiverseSD}, and presented in \autoref{eq:alg_heuristic_cover},
\begin{equation}
\label{eq:alg_heuristic_cover}
\eta_{W}(I_{ij}) = \frac{1}{|G_{I_{ij}}|} \sum\limits_{o \in G_{I_{ij}}} W^{g(o, \mathbb{G})}
\end{equation}
where $G_{I_{ij}}$ is the subgroup defined by the description $D(o) = A_i(o)\in\{v_{ij}\}$, $g(o, \mathbb{G}) = |\{G \in \mathbb{G}| o \in G\}|$ is the number of subgroups in $\mathbb{G}$ that contain an observation $o$, and $W \in (0,1]$ is the weight parameter.
Hence, the less often the observations described by $I_{ij}$ are already covered by subgroups in $\mathbb{G}$, more likely it is for the item to be visited in future iterations of the algorithm.

It is important to notice that $\eta_{L}$ takes into consideration the descriptions of all subgroups already discovered by the algorithm -- whether or not they are included in $\mathbb{G}$.
In the other hand, $\eta_{W}$ only considers the coverage of the subgroups currently in $\mathbb{G}$.

\subsection{Subgroup Search}
\begin{algorithm}[t]
    \scriptsize
    \caption{Subgroup Search}
    \label{alg:esmam_subgroupsearch}
    \DontPrintSemicolon
      
    \KwInput{$\mathcal{B}$ -- baseline for subgroup comparison, $nAnts$ -- size of the ant colony, $nConverg$ -- number of similar patters for convergence, $minCov$ -- minimum subgroup coverage}
    \KwOutput{$G^{best}$ -- discovered subgroup}
    
    \SetKwFunction{Fsearch}{subgroupSearch}
    \SetKwProg{Fn}{Function}{:}{}
    \Fn{\Fsearch{$\mathcal{B}, nAnts,nConverg,minCov$}}{
    
    $ant \leftarrow 0$; $converg \leftarrow 0$\\
    $G^- \leftarrow \varnothing$; $G^{best} \leftarrow \varnothing$\\
    
   	\While{$t \leq nAnts$ \textbf{or} $converg \leq nConverg$}
    {
        $D \leftarrow$ \texttt{buildDescription}$(minCov)$\\
        $D \leftarrow$ \texttt{pruneDescription}$(D, \mathcal{B})$\\
        \texttt{pheromoneUpdating}$(\mathcal{I}(G_D))$\\
        
        \If{$\phi(G_D, \mathcal{B}) > \phi(G^{best}, \mathcal{B})$}
        {
            $G^{best} \leftarrow G_D$
        }
        \If{$\mathcal{I}(G_D) = \mathcal{I}(G^-)$}{
       	    $converg \leftarrow converg + 1$
       	}
       	\textbf{else:} $converg \leftarrow 0$\\
        $G^- \leftarrow G_D$\\
        $t \leftarrow t + 1$\\
        
    }

    \textbf{return:} $G^{best}$
    } 
\end{algorithm}

In the EsmamDS framework (\autoref{alg:esmam_framework}), once the pheromone and heuristic values are initialised (line 5), the \texttt{subgroupSearch} function implements the ACO search and returns a single subgroup (line 6).
The pseudocode of the function is presented in \autoref{alg:esmam_subgroupsearch}.

Each ant $t$ in a colony of $nAnts$ ants delivers a complete description $D$ in a two-step process: the stochastic description construction (line 5); and a local search pruning procedure (line 6).
Then, the items $\mathcal{I}(G_D)$ represented in the constructed description are used to update the pheromone trail for the next ant.
This process is repeated for all ants in the colony, or until the ants converge to a solution (line 4) -- i.e. until the colony achieves a minimum threshold ($nConverg$) for identical sequential descriptions (lines 10-12).
The best subgroup $G^{best}$ (according to the quality measure $\phi$) discovered within the colony is, then, returned.

The \texttt{buildDescription} (line 5) is a refinement function that starts from an empty partial description $D^p$ and iteratively generates a more complex description by adding conjunctive conditions $A_i\in \{v_{ij}\}$ to $D^p$; note there is only a single constraint on each attribute.
The addition of such conditions is a stochastic procedure that chooses items considering both their heuristic and pheromone values. 
They are sorted from the set of items belonging to the observations covered by $D^p$ with the probability given in \autoref{eq:alg_probability},
\begin{equation}
\label{eq:alg_probability}
\forall I_{ij} \in \gamma(G_{D^p}) \left( P(I_{ij}) = \frac{x_i \cdot \eta(I_{ij}) \cdot \tau^t(I_{ij})}{\sum\limits_{I_{ij}} x_i \cdot  \eta(I_{ij}) \cdot \tau^t(I_{ij})} \right)
\end{equation}
where $x_i = 1$ if $A_i$ is not yet represented in $D^p$, and zero otherwise.
This refinement process of iteratively sorting items stops when all $A_i$ are represented in $D^p$ or when a new condition results in a coverage $|G_{D^p}|$ below a minimum threshold $minCov$.
Note that the final constructed description $D$ is a conjunction of conditions over singleton sets of $dom(A_i)$.

When a full description $D$ is constructed, a generalisation function \texttt{pruneDescription} (line 6) is responsible for enhancing both the simplicity and the quality of the final solution.
This procedure greedly removes conditions $A_i\in\{v_{ij}\}$ from $D$, each time eliminating the condition that leads to the largest improvement in the quality associated with the (pruned) description.
The pruning stops when no conditions can be removed without decreasing the quality or when the description already encompasses only a single condition.

Finally, the \texttt{pheromoneUpdating} function (line 7) generates the pheromone values $\tau^{t+1}$ for the next ant iteration.
For the items represented in the final description $D$, the pheromone is incremented proportionally to the subgroup's quality, as given bellow.
$$\tau^{t+1}(I_{ij}) = \tau^t(I_{ij}) + \phi(G_D,\mathcal{B}) \cdot \tau^{t}(I_{ij}), \;iif\; I_{ij} \in \mathcal{I}(G_D)$$
For the items not represented in $D$, an evaporation process is simulated by the normalisation of all $\tau$ values in $(t+1)$.
\subsection{Subgroup Set Updating}

In each iteration of the EsmamDS algorithm (\autoref{alg:esmam_framework}), after a subgroup $G$ is discovered (line 6), the final set of subgroups $\mathbb{G}$ is updated considering the inclusion of $G$.
The \texttt{subgroupSetUpdating} (line 7) method is a recursive function that adjusts the subgroup set to (i) minimise both descriptive and model redundancies, and (ii) maximise the coverage of subgroups, i.e. improve their generalisation.

By allowing descriptions to constrain attributes on a set of values, generalisation is achieved since an extended description may subsume a set of subgroups.
Hence, we first introduce two generalisation operations between subgroups that combine two descriptions into a single subgroup: (i) the \emph{root} operator, that provides a common generalisation between two descriptions; and (ii) the \emph{merge} operator, that unifies two different descriptions into a more general one.
Lets consider the three descriptions as follows:
\begin{table}[H]
\begin{center}
\begin{tabular}{l l l l}
$D_1:$  & $A_i \in \{v_{i1}\}$  & $\land$   & $A_j \in \{v_{j1},v_{j2}\}$ \\
$D_2:$  & $A_i \in \{v_{i1},v_{i2}\}$  & $\land$   & $A_j \in \{v_{j2},v_{j3}\}$\\
$D_3:$  & $A_i \in \{v_{i3}\}$  & $\land$   & $A_k \in \{v_{k1}\}$
\end{tabular}
\end{center}
\end{table}
Considering the pair $(D_1,D_2)$, the \emph{root} operation yield a new description $D_r: A_i \in \{v_{i1}\} \land A_j \in \{v_{j2}\}$.
Thus, when employing the \emph{merge} operation, we have the generalisation $D_m: A_i \in \{v_{i1},v_{i2}\} \land A_j \in \{v_{j1},v_{j2},v_{j3}\}$.
Note, however, that the description $D_3$ does not have a \emph{root} with neither $D_1$ nor $D_2$ because they do not present a common attribute constrain.
Additionally, it also cannot be merged with neither $D_1$ nor $D_2$ because they constrain different attributes.

Formally, we have that, given two subgroups $G_a$ and $G_b$:
\begin{itemize}
    \item $root(G_a, G_b) = \mathcal{I}(G_a) \cap \mathcal{I}(G_b)$ provided that the intersection exists; and
    \item $merge(G_a, G_b) = \mathcal{I}(G_a) \cup \mathcal{I}(G_b)$ if and only if the attributes $A_i$ represented in both $G_a$ and $G_b$ descriptions are exactly the same.
\end{itemize}
Additionally, we define that $G_a$ \emph{is-in} $G_b$ if $\mathcal{I}(G_a) \subseteq \mathcal{I}(G_b)$ given that the attributes $A_i$ represented in both their descriptions are exactly the same.
In this case, we can also say that $G_b$ is a generalisation of $G_a$.

Finally, the pseudocode of the \texttt{subgroupSetUpdating} function is provided in \autoref{alg:updateSgSet}.
\begin{algorithm}[!t]
    \scriptsize
    \caption{Subgroup Set Updating}
    \label{alg:updateSgSet}
    \DontPrintSemicolon
      
    \KwInput{$G_{new}$ -- subgroup, $\mathbb{G}$ -- set of subgroups, $\alpha$ -- level of significance}
    \KwOutput{$\mathbb{G}$ -- set of exceptional subgroups}
    
    \SetKwFunction{Fset}{subgroupSetUpdating}
    \SetKwProg{Fn}{Function}{:}{}
    \Fn{\Fset{$G_{new}, \mathbb{G}, \alpha$}}{
    
    \If{$\phi(G_{new}, \mathcal{B}) < 1 - \alpha$:}{
         \textbf{return:} $\mathbb{G}$
    }
    
    \For{$G \in \mathbb{G}$}
    {   
        \If{$G_{new}, G$ have different models \textbf{or} strictly different attributes}{
            \textbf{next}
        }
        
        \Else
        {
            \If{ $G_{new}$ \emph{is-in} $G$:}{
                 \textbf{return:} $\mathbb{G}$
            }
            
            \If{$G$ \emph{is-in} $G_{new}$}
            {
                $\mathbb{G}' \leftarrow$ \texttt{subgroupSetUpdating}$(G_{new}, \mathbb{G}\setminus G, \alpha)$\\
                \textbf{if} $G_{new} \in \mathbb{G}'$: \textbf{return:} $\mathbb{G}'$\\
                \textbf{else}: \textbf{return:} $\mathbb{G}$\\
            }
            

            \If{\textbf{not} $root(G_{new},G)$ nor $merge(G_{new},G)$}
            {
                \textbf{next}
            }
            
            \Else
            {   
                $G_r \leftarrow root(G_{new},G)$\\
                $G_m \leftarrow merge(G_{new},G)$\\
                $\mathbb{G}' \leftarrow$ \texttt{subgroupSetUpdating}$(G_r, \mathbb{G}, \alpha)$\\
                $\mathbb{G}' \leftarrow$ \texttt{subgroupSetUpdating}$(G_m, \mathbb{G}', \alpha)$\\
                
                \textbf{if} \emph{neither} $G_r,G_m \in \mathbb{G}'$: \textbf{next}\\
                
                \If{only $G_r \in \mathbb{G}'$}
                {
                    \textbf{if} $(G_r,G_{new})$ models are different: \textbf{next}\\
                    \textbf{else}: \textbf{return:} $\mathbb{G}'$\\
                }
                \If{only $G_m \in \mathbb{G}'$}
                {
                    \textbf{if} $(G_m,G_{new})$ models are different: \textbf{next}\\
                    \textbf{else}: \textbf{return:} $\mathbb{G}'$\\
                }
                \If{both $G_r,G_m \in \mathbb{G}'$}
                {
                    \textbf{if} $G_{new}$ model differs from both $G_r, G_m$: \textbf{next}\\
                    \textbf{else}: \textbf{return:} $\mathbb{G}'$\\
                }
            }

        }
    }

    \textbf{return:} $\mathbb{G} \leftarrow \mathbb{G} \cup \{G_{new}\}$
    } 
\end{algorithm}
A new candidate subgroup $G_{new}$ is added to the set $\mathbb{G}$ if it satisfies a lower quality bound for assuring exceptionality (line 2) \textbf{and} if (for all subgroups $G \in \mathbb{G}$) (lines 4-6):
\begin{itemize}
    \item $G_{new}$ and $G$ have statistically different survival models; \textbf{or}
    \item $G_{new}$ represents only attributes $A_i$ not represented in $G$.
\end{itemize}
In other words, we select the exceptional subgroups that present a previously unobserved behaviour or a completely new characterisation.
We compare the survival models of two subgroups with the log-rank test considering a level of significance of $\alpha$.

Hence, we further process the cases where a pair $G_{new},G$ present similar models and some resemblance in description (lines 7-30).
In those cases, we first assess whether there is a more general subgroup to keep using the \emph{is-in} comparison (lines 8-13).
Otherwise, we employ the \emph{root} and \emph{merge} operators (when possible) to obtain subgroups' generalisations (lines 14-30).
%
%
%
\section{Experimental Procedure}
\label{sec:experiments}

We conducted experiments to evaluate our proposed approach for mining local patterns associated with exceptional survival behaviours.
We aim at providing comprehensible and diverse characterisation of subgroups presenting unusual KM models.
We compare our algorithm with state of the art methods in the literature that provide characterisation over unusual survival behaviour: (i) the Esmam algorithm \cite{my:bracis}, an ACO heuristic approach to mine unusual survival models (which served as base for this work); (ii) the beam-search heuristic for mining subgroups, considering either a single target (SD) and a model target (EMM); (iii) the \emph{DSSD-CBSS} algorithm \citep{SD:algorithm_DiverseSD}, a beam-search approach to mine a diverse set of subgroups adapted to use a similar target as ours; and (iv) the \emph{LR-Rules} \citep{SA:article_EMM}, a covering rule-base algorithm for predicting survival response (although the ultimate goal of this method is to build a predictive model, we decided to include it in our study as its authors also suggested its application for finding descriptions). 
We assess the results with respect to the exceptionality, interpretability, generality and redundancy of the findings.
\autoref{tbl:eval_metrics} presents all metrics used in the results evaluation.
We consider both exceptionality and generality maximisation metrics; for the others, we consider minimisation.
\begin{table}[t!]
\begin{center}
\caption{Evaluation metrics}
\label{tbl:eval_metrics}
\scalebox{0.90}{
\begin{tabular}{p{0.5cm} p{2.2cm} >{\centering\arraybackslash}p{6cm}}
\hline
\makecell[cl]{\textbf{Metrics}} &     \makecell[cc]{\textbf{Description}} &  \makecell[cc]{\textbf{Definition}} \\
\hline
\multicolumn{3}{l}{\textbf{Similarity}}\\
\makecell[tl]{$sim_D$}   & \makecell[tl]{Description \\similarity}    & $sim_D(G_a,G_b) = \frac{|\mathcal{I}(G_a) \cap \mathcal{I}(G_b)|}{min(|\mathcal{I}(G_a)|,|\mathcal{I}(G_b)|)}$\\
\makecell[tl]{$sim_C$}   & \makecell[tl]{Coverage\\similarity}       & $sim_C(G_a,G_b) = \frac{|G_a \cap G_b|}{min(|G_a|,|G_b|)}$\\
\makecell[tl]{$sim_M$}   & \makecell[tl]{Model\\similarity}          & $sim_M(G_a,G_b) = p_{KM} > \alpha$\\
\hline
\multicolumn{3}{l}{\textbf{Exceptionality}}\\
\makecell[tl]{$\epsilon$}           & \makecell[tl]{Proportion of exceptional\\subgroups}        & $\frac{\sum\limits_{G \in \mathbb{G}} sim_M(G,\mathcal{B})}{|\mathbb{G}|}$\\
\hline
\multicolumn{3}{l}{\textbf{Interpretability}}\\
\makecell[tl]{\emph{\#sg}}        & \makecell[tl]{Number of discovered\\subgroups}        & $|\mathbb{G}|$\\
\makecell[tl]{\emph{lenght}}      & \makecell[tl]{Average subgroup\\description length}   & $\frac{\sum\limits_{G \in \mathbb{G}} size(G)}{|\mathbb{G}|}$\\
\hline
\multicolumn{3}{l}{\textbf{Generality}}\\
\makecell[tl]{\emph{sgCov}}       & \makecell[tl]{Average (percentage)\\subgroup coverage}  & $\frac{\sum\limits_{G \in \mathbb{G}} |G|}{|\mathbb{G}| \cdot |\Omega|} $\\
\makecell[tl]{\emph{dbCov}}       & data set coverage           & $\frac{|\bigcup\limits_{G \in \mathbb{G}} G|}{|\Omega|} $\\
\hline
\multicolumn{3}{l}{\textbf{Redundancy}}\\
\makecell[tl]{$\rho_D$}   & \makecell[tl]{Description \\redundancy}    & $\rho_D = \binom{\mathbb{G}}{2}^{-1} \sum\limits_{G_a, G_b \in \mathbb{G}, G_a \neq G_b} sim_D(G_a, G_b)$\\
\makecell[tl]{$\rho_C$}   & \makecell[tl]{Coverage\\redundancy}       & $\rho_C = \binom{\mathbb{G}}{2}^{-1} \sum\limits_{G_a, G_b \in \mathbb{G}, G_a \neq G_b} sim_C(G_a, G_b)$\\
\makecell[tl]{$\rho_M$}   & \makecell[tl]{Model\\redundancy}          & $\rho_M = \binom{\mathbb{G}}{2}^{-1} \sum\limits_{G_a, G_b \in \mathbb{G}, G_a \neq G_b} sim_M(G_a, G_b)$\\
\makecell[tl]{$CR$}                 & Cover Redundancy          & \autoref{eq:cr}\\
\hline
\end{tabular}}
\end{center}
\end{table}

The similarity metrics are comparisons between pairs of subgroups (note that the baseline $\mathcal{B}$ can be considered a subgroup to compare with).
The \emph{description} similarity $sim_D: \mathcal{P}(\mathbb{I}) \rightarrow [0,1]$ and the \emph{coverage} similarity $sim_{C}: \mathcal{P}(\Omega) \rightarrow [0,1]$ are an adaptation of the Jaccard index more sensible to the overlaps between the compared sets.
The \emph{model} similarity $sim_M$ is assessed as a boolean function based on the log-rank test, where $p_{KM}$ is the $p$-value of the test between the models of the two compared subgroups and $\alpha$ is a predefined level of significance.
By  contrast, the metrics of exceptionality, interpretability, generality and redundancy are global metrics for a \emph{set of subgroups}.

Redundancy is assessed -- for descriptions, coverage, and survival models -- as the normalised sum of the similarity measures for all (unordered) pairs of subgroups in the set.
Additionally, we also evaluate coverage redundancy using the CR measure \cite{SD:algorithm_DiverseSD}.
Such measure quantifies the extent of the deviation between the coverage of the subgroups in a set $\mathbb{G}$ from a uniform (cover) distribution.
Being $g(o, \mathbb{G})$ the number of subgroups in $\mathbb{G}$ that cover an observation $o$, we have that the expected number of times for a random observation to be covered is $\hat{g} = |\mathbb{G}|^{-1}\sum_{o \in \Omega}g(o, \mathbb{G})$.
Then, the CR is defined as presented in \autoref{eq:cr}.
\begin{equation}
    \label{eq:cr}
    CR = \frac{1}{|\Omega|}\sum_{o\in \Omega}\frac{|g(o,\mathbb{G}) - \hat{g}|}{\hat{g}}
\end{equation}
High values of this measure indicate that the observations contained in the subgroups of $\mathbb{G}$ are covered more than expected. 
In other words, a large number of subgroups in the set cover the same observations. 
Hence, low values of CR indicate more diversity/less redundancy between subgroups.

Next, we describe the process of empirical evaluation and analyse the results.
Some contents like EsmamDS implementation, the data sets used in the tests, configurations and results are available on \href{https://github.com/jbmattos/EsmamDS/tree/icde2022}{EsmamDS website}.
%
%
\subsection{Experimental Setup}
\label{subsec:experimental_setting}

We conduct experiments with 14 real-world survival data sets from the medical domain. 
These data sets were used as benchmark data in
many survival analysis studies. 
Besides being frequently used as benchmark data, they are also of particular interest for us since we
believe that one of the most suitable applications for our method is in the medical domain. 
In this case, we can also observe the behaviour of our approach in this domain. 
\autoref{tbl:db-Description14} contains the list of these data sets together with a brief characterisation.

The patterns we analyse are searched in the space of the domains of attributes $A_i$  (i.e. the set of items $\mathbb{I}$).
Then, the survival behaviour is analysed through the deviation of the target concept over the target features $T, \delta$.
As the EsmamDS algorithm is not adapted to process high-dimensional data, we selected data sets of low dimensionality.
All numerical descriptive attributes were discretised using equal-frequency discretisation into five interval categories. 
Pre-processing of the data was employed to remove observations containing missing values (and features with a high level of missing data).
All data sets and their general information can be found in the EsmamDS \href{https://github.com/jbmattos/EsmamDS/tree/main/data\%20sets}{repository}.
\begin{table}[t]
\begin{center}
\caption{Characteristics of 14 survival data sets used in the experimental study: the number of observations ($|\Omega|$), the number of descriptive attributes ($|A_i|$), the number of descriptive attributes $A_i$ that were discretised ($|A_i^d|$), the number of items in the data set ($|\mathbb{I}|$), the proportion of censored observations ($\%cens$), the \emph{Subject of Research}, and the survival event description ($Event$)}
\label{tbl:db-Description14}
\scalebox{0.90}{
\begin{tabular}%
    {>{\raggedright\arraybackslash}p{1.5cm} %
    >{\raggedleft\arraybackslash}p{0.3cm} %
    >{\raggedleft\arraybackslash}p{0.2cm} %
    >{\raggedleft\arraybackslash}p{0.3cm} %
    >{\raggedleft\arraybackslash}p{0.2cm}%
    >{\raggedleft\arraybackslash}p{0.6cm}%
    p{1.8cm}%
    p{1.4cm}}
\hline
data set ($\Omega$) & $|\Omega|$ & $|A_i|$ & $|A_i^d|$    & $|\mathbb{I}|$    & \%cens    &  \makecell[cl]{Subject of\\research}     & Event \\
\hline
actg320           &  1151 &     11 &    3       & 39      &   91.66    &  \makecell[tl]{HIV-infected\\patients}        & \makecell[tl]{AIDS death/\\diagnosis}     \\
breast-cancer     &   196 &     80 &    78      & 269     &   73.98    &  Node-Negative breast cancer                  & \makecell[tl]{distant\\metastasis}        \\
cancer            &   168 &      7 &    5       & 29      &  27.98    &  \makecell[tl]{Advanced\\lung cancer}         & death                                     \\
carcinoma         &   193 &      8 &    1       & 28      &  27.46    &  \makecell[tl]{Carcinoma of\\the oropharynx}  & death                                     \\
gbsg2             &   686 &      8 &    5       & 31      &  56.41    &  \makecell[tl]{Breast\\cancer}                & recurrence                                \\
lung              &   901 &      8 &    0       & 23      &  37.40    &  \makecell[tl]{Early lung\\cancer}            & death                                     \\
melanoma          &   205 &      5 &    3       & 28      &  72.20    &  Malignant melanoma                           & death                                     \\
mgus              &   176 &      8 &    6       & 30      &   6.25    &  \makecell[tl]{Monoclonal\\gammopathy}        & death                                     \\
mgus2             &  1338 &      7 &    5       & 23      &  29.90    &  \makecell[tl]{Monoclonal\\gammopathy}        & death                                     \\
pbc               &   276 &     17 &    10      & 61      &  59.78    &  \makecell[tl]{Primary biliary\\cirrhosis}    & death                                     \\
ptc               &   309 &     18 &    1       & 71      &  93.53    &  \makecell[tl]{Papillary thyroid\\carcinoma}  & \makecell[tl]{recurrence/\\progression}   \\
uis               &   575 &      9 &    4       & 33      &  19.30    &  \makecell[tl]{Drug addiction\\treatment}     & \makecell[tl]{return to\\drug use}        \\
veteran           &   137 &      6 &    3       & 23      &   6.57    &  \makecell[tl]{Lung\\cancer}                  & death                                     \\
whas500           &   500 &     14 &    6       & 46      &  57.00    &  \makecell[tl]{Worcester\\ Heart Attack}      & death                                     \\
\hline
\end{tabular}}
\end{center}
\end{table}

We conducted experiments to assess the performance of the EsmamDS considering both baselines $\mathcal{B}$ for subgroup comparison, \emph{population} and \emph{complement}, and evaluated them separately.
In the empirical evaluation, we performed 30 executions for each data set due to its stochastic nature. 
The proper configuration of the algorithm was defined with a randomised search considering the following parameters' values: $nAnts = \{100, 200, 500, 1000, 3000\}$; $minCov = \{0.01, 0.02, 0.05, 0.1\}$; $nConverg = \{5, 10, 30\}$; $maxStag = \{20, 30, 40, 50\}$; $L = \{1,3,5,10\}$; $W = 0.9$ (according to employed in \cite{SD:algorithm_DiverseSD}); and $\alpha = 0.05$.
We sampled 10\% of the total number of combinations and, then, executed the EsmamDS for three data sets (namely: actg320, breast-cancer and ptc).
The best configuration (for each baseline) was chosen by ordering all configuration samples from the random search according to their average performance for the following metrics' order: $\rho_\mathcal{D}$, $\rho_\mathcal{C}$, $CR$, $\rho_\mathcal{M}$, $sgCov$, $dbCov$, $length$ and $\#sg$.

For the configuration of the other approaches, for each data set and according to a baseline, we used the results achieved by the EsmamDS in the empirical evaluation to adjust the following three parameters:
\begin{itemize}
    \item ($minCov$) Minimum coverage: defined by the same parameter value chosen for the EsmamDS;
    \item ($bs$) Beam-size (or maximum number of discovered subgroups): given by the average number of subgroups discovered by the EsmamDS in the 30 experiments;
    \item ($maxDepth$) Rule-depth (or refinement/search depth): given by the average of the the maximum description size achieved during the EsmamDS execution in the 30 experiments.
\end{itemize}

\autoref{tbl:algorithms} displays the configuration for each baseline of all the algorithms compared.
For the beam-search approaches that consider single target, we employed a quality measure given as $1 - p_{T}$, being $p_{T}$ the \emph{p-value} of the bilateral t-Test for the survival time.
For the DSSD-CBSS algorithm, the quality measure was defined as the t-Test, and the remaining approaches employ the measure defined in \autoref{eq:quality_emm}.
\begin{table*}[t]
\begin{center}
\caption{Information on the algorithms compared in this work}
\label{tbl:algorithms}
\scalebox{0.88}{
\begin{tabular}{p{.4em}p{1.8cm} p{1.6cm} p{1.54cm} p{3cm} p{10cm}}
\hline
&\makecell[cl]{Algorithm} &     \makecell[cl]{Search\\Strategy} &   \makecell[cl]{Target\\Concept} &   \makecell[cl]{Source [Reference]} &   \makecell[cl]{Parameter(Value)} \\
\hline
\multirow{4}{*}{\rotatebox[origin=c]{90}{\hspace{0.3em}\textbf{Population \hspace{0.3em}}}}&
EsmamDS-pop &     ACO           &   KM model        & \href{https://github.com/jbmattos}{EsmamDS repository} & $\alpha(0.05)$, $nAnts(100)$, $minCov(0.1)$, $nConverg(5)$, $maxStag(40)$, $W(0.9)$, $L(5)$\\
&
Esmam-pop   &     ACO           &   KM model        & \href{https://github.com/jbmattos/ESM-AM_bracis2020}{Esmam repository}  \cite{my:bracis} & $\alpha(0.05)$, $nAnts(100)$, $minCov(0.1)$, $nConverg(5)$, $maxStag(40)$\\
&
BS-EMM-pop  &     \makecell[tl]{Beam Search}   &   KM model        & \href{https://github.com/flemmerich/pysubgroup}{PySubgroup package} \cite{EMM:framework_Pysubgroup} & $minCov(0.1)$, $bs$, $maxDepth$\\
&
BS-SD-pop   &     \makecell[tl]{Beam Search}   &   Survival time   & PySubgroup package & $minCov(0.1)$, $bs$, $maxDepth$\\ 
&
DSSD-CBSS   &     \makecell[tl]{Beam Search} & Survival time & \href{https://datamining.liacs.nl/cortana.html}{Cortana package} \cite{SD:algorithm_DiverseSD} & \emph{search strategy(Cover-based beam selection)}, $minCov(0.1)$, $bs$, $maxDepth$, $time(\infty)$\\
\hline
\vspace{0.5em}
\multirow{4}{*}{\rotatebox[origin=b]{90}{\textbf{Complement}}}&
EsmamDS-cpm &     ACO           &   KM model        & EsmamDS repository & $\alpha(0.05)$, $nAnts$ (100), $minCov(0.05)$, $nConverg(5)$, $maxStag(40)$, $W(0.9)$, $L(10)$\\
&
Esmam-cpm   &     ACO           &   KM model        & Esmam repository  & $\alpha(0.05)$, $nAnts(100)$, $minCov(0.05)$, $nConverg(5)$, $maxStag(40)$\\
&
BS-EMM-cpm  &     \makecell[tl]{Beam Search}   &   KM model        & PySubgroup package & $minCov(0.05)$, $bs$, $maxDepth$\\
&
BS-SD-cpm   &     \makecell[tl]{Beam Search}   &   Survival time   & PySubgroup package & $minCov(0.05)$, $bs$, $maxDepth$\\
&
LR-Rules    &     \makecell[tl]{Sequential\\covering} & KM model    & \href{https://github.com/adaa-polsl/LR-Rules/releases}{LR-Rules repository} \cite{SA:article_EMM} & --\\
\hline
\end{tabular}}
\end{center}
The $bs, maxDepth$ parameters were configured  for each data set as defined in the text. The remaining user-defined configuration of the frameworks were kept as default.
The specifics of all configurations are provided \href{https://github.com/jbmattos/EsmamDS/blob/icde2022/experiments/_paramsConfig.csv}{here}.
\end{table*}

Finally, statistical analysis of the results was performed by a (paired) Friedman test followed by a Nemenyi post-hoc test.
We performed the Friedman test to compare whether or not the compared approaches present statistically similar performances (null hypothesis) regarding the proposed metrics.
When the Friedman's null hypothesis is rejected, we proceed with the Nemenyi test to validate which approaches stand out in their performances.
We executed the tests using EsmamDS (and Esmam) complete sample of 420 results for each metric (30 experiments on 14 data sets).
For the remaining deterministic algorithms, we paired the results for each data set by repeating them 30 times.
We assessed the tests using a level of significance of $5\%$.
%
%
\subsection{Results Analysis}
\label{subsec:results_analysis}

\begin{table*}[t]
\begin{center}
\caption{Evaluation metrics computed over the results provided by the compared approaches: the metrics' average over all data sets (Avg.); and the mean rank of the Nemenyi post-hoc test computed for each metric (Rank).}
\label{tbl:results}
\scalebox{0.9}{
\begin{tabular}{p{.4em}l c c c c c c c c c c c c c c c c c}
\hline
&\multirow{2}{*}{\textbf{Algorithms}} & $\epsilon$ & \multicolumn{2}{c}{$\#sg$} & \multicolumn{2}{c}{$length$} & \multicolumn{2}{c}{$sgCov$} & \multicolumn{2}{c}{$setCov$} & \multicolumn{2}{c}{$\rho_D$} & \multicolumn{2}{c}{$\rho_C$} & \multicolumn{2}{c}{$CR$} & \multicolumn{2}{c}{$\rho_M$} \\
&           &   Avg.   &   Avg. & Rank &   Avg. & Rank &    Avg. & Rank &   Avg. & Rank &                Avg. & Rank &             Avg. & Rank & Avg. & Rank &             Avg. & Rank \\
\hline
\multirow{4}{*}{\rotatebox[origin=b]{90}{\hspace{1cm}\textbf{Population}}}&
EsmamDS-pop & \textbf{1.00} & \textbf{5.05} & \textbf{2.28} & 1.52          & 1.93          & \textbf{0.28} & \textbf{1.37} & \textbf{0.87} & \textbf{1.14} & \textbf{0.03} & \textbf{1.23} & \textbf{0.23} & \textbf{1.25} & \textbf{0.31} & \textbf{1.09} & \textbf{0.26} & \textbf{1.22} \\
& Esmam-pop   & \textbf{1.00} & 7.31          & 3.51          & \textbf{1.46} & \textbf{1.74} & 0.21          & 2.33          & 0.64          & 2.25          & 0.17          & 2.11          & 0.39          & 2.39          & 0.51          & 2.39          & 0.56          & 2.40          \\
& BS-EMM-pop  & \textbf{1.00} & 5.50          & 3.07          & 2.22          & 3.57          & 0.16          & 4.12          & 0.33          & 4.14          & 0.54          & 3.86          & 0.67          & 4.07          & 0.70          & 4.27          & 0.84          & 3.91          \\
& BS-SD-pop   & 0.81          & 5.50          & 3.07          & 2.46          & 3.33          & 0.17          & 3.39          & 0.39          & 3.40          & 0.46          & 3.34          & 0.54          & 3.33          & 0.62          & 3.21          & 0.69          & 3.10          \\
& DSSD-CBSS   & 0.71          & 5.50          & 3.07          & 2.63          & 4.43          & 0.17          & 3.79          & 0.27          & 4.07          & 0.69          & 4.46          & 0.73          & 3.95          & 0.74          & 4.05          & 1.00          & 4.36          \\
\hline
\multirow{4}{*}{\rotatebox[origin=b]{90}{\textbf{Complement}}}&
EsmamDS-cpm & \textbf{1.00} & \textbf{5.36} & \textbf{2.31} & \textbf{1.22} & \textbf{1.78} & \textbf{0.34} & \textbf{1.57} & 0.98          & 2.13          & \textbf{0.02} & \textbf{1.54} & 0.33          & 2.33          & \textbf{0.29} & \textbf{1.78} & \textbf{0.24} & \textbf{1.84} \\
& Esmam-cpm   & \textbf{1.00} & 5.70          & 2.58          & 1.27          & 1.84          & 0.25          & 2.92          & 0.71          & 3.18          & 0.10          & 2.23          & 0.40          & 2.95          & 0.46          & 3.28          & 0.45          & 3.20          \\
& BS-EMM-cpm  & \textbf{1.00} & 6.00          & 3.10          & 2.16          & 3.98          & 0.19          & 4.08          & 0.50          & 4.00          & 0.51          & 4.37          & 0.67          & 4.30          & 0.63          & 4.05          & 0.66          & 4.19          \\
& BS-SD-cpm   & 0.82          & 6.00          & 3.10          & 2.56          & 4.29          & 0.18          & 3.93          & 0.50          & 4.10          & 0.41          & 4.07          & 0.49          & 3.50          & 0.59          & 3.90          & 0.52          & 3.35          \\
& LR-Rules    & 0.94          & 9.43          & 3.91          & 1.86          & 3.11          & 0.28          & 2.50          & \textbf{1.00} & \textbf{1.59} & 0.15          & 2.66          & \textbf{0.32} & \textbf{1.84} & 0.32          & 1.99          & 0.30          & 2.34          \\
\hline
\end{tabular}}
\end{center}
EsmamDS and Esmam results were averaged over 30 experiments for each data set.
\end{table*}

In this section, we present and analyse the results achieved by the algorithms.
\autoref{tbl:results} contains the average performance for all evaluation metrics, except the similarity ones (that will be approached later in this section).
We performed the Friedman test for the metrics of interpretability, generality and redundancy.
For all tested metrics, we rejected the null hypothesis that the  algorithms present similar performances.
Hence, we employed the Nemenyi post-hoc test for each metric to assess the differences between the performances, The average rank used in the Nemenyi test for each metric is also provided in the table. We now discuss each of these metrics. 

The $\epsilon$ metric of exceptionality reveals the unusualness of the survival behaviour associated with the discovered subgroups.
We inspected each subgroup to verify if its survival model is exceptional (statistically different) when compared to the survival model of the considered baseline.
We observe that the methods that employ the subgroup discovery approach rather than exceptional model mining, namely BS-SD-pop, DSSD-CBSS and BS-SD-cpm, do not assure the discovery of exceptional models.
This is expected because the occurrence of outliers in a subgroup may significantly bias its average survival time, while not necessarily changing the event distribution over time.
In addition, the predictive sequential covering LR-Rules also delivers some subgroups that are not unusual.
Although those approaches provide insights on divergences in survivability, when aiming to characterise survival behaviours (models), it is crucial to properly represent the target concept to be optimised. 
In this sense, the EMM approach plays an important role in behavioural analysis.

When assessing the statistical results for the Friedman and Nemenyi tests, we notice that the EsmamDS (for both baselines) outperforms all other approaches in $sgCov$ -- presenting an average coverage from 7 to 16\% higher than the others -- and in $\rho_D$ and $\rho_M$ with a reduction of 80-95\% in the levels of descriptive redundancy and 20-74\% in the levels of model redundancy.
The EsmamDS(-\emph{cpm}) is outperformed only by the covering LR-Rules in $setCov$ (a difference of 2\%) and in the coverage redundancy $\rho_C$ (a difference of 0.01 in the metric average), but being similar in the $CR$ performance. 
It is important to notice that the LR-Rules stopping criteria includes
covering all the individuals in the data set, which justifies the difference in coverage. 
Also it naturally induces rules for disjoint subsets of individuals. 
At each turn a new rule have to cover a minimum number of previously uncovered individuals from the complete data set.
This justifies the low redundancy in coverage.
By contrast, the EsmamDS-\emph{pop} outperforms all other approaches (the beam search algorithms) in those same metrics.
We observe a (statistically) significant reduction of 31\% in $\#sg$ when comparing EsmamDS-pop to Esmam-pop; 43\% comparing EsmamDS-cpm to the LR-Rules; and 6\% comparing to Esmam-cpm (with no statistical difference).
Furthermore, no statistical difference was observed for the $length$ metric between EsmamDS and its predecessor Esmam.

\begin{figure*}[t!]
\centering
\includegraphics[width=\textwidth]{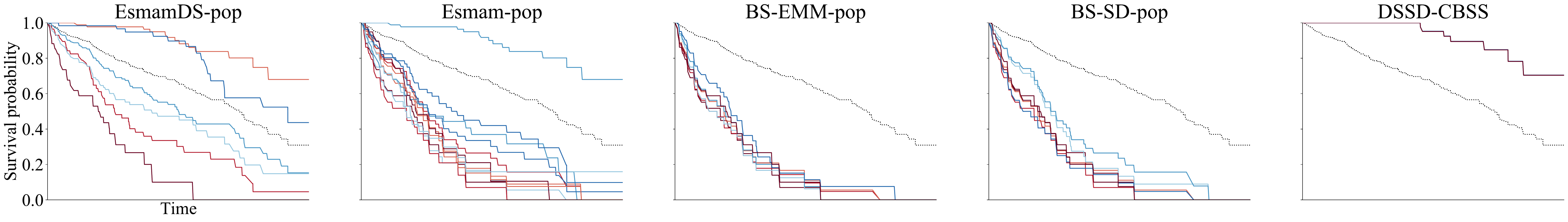}
\label{subfig:surv_pop_1}
\caption{Survival curves of the subgroups discovered by the $\mathcal{B}$-\emph{population} algorithms on \emph{pbc} data set (experience 0). The dotted line is the survival curve fitted on the data set.}
\label{fig:surv_pop}
\end{figure*}

Hence, our approach delivers small sets of patterns ($\#sg$) that comprise more compact characterisations ($length$) while providing more generalisation ($sgCov$).
It was capable of characterising the large majority of the data records ($setCov$) and consistently achieves lower redundancy levels (in all three dimensions).
Thus, it provides subgroups that present lower similarity in description ($\rho_D$) and coverage ($\rho_C, CR$) while delivering a larger variety of exceptional survival models ($\rho_M$).

Regarding survival behaviours (models), the model redundancy metric $\rho_M$ shows that our approach attains higher diversity by achieving lower proportions of similar (pair of) models.
Such a result is endorsed by the KM plots of the discovered subgroups, presented in \autoref{fig:surv_pop}.
Throughout the plots, it is possible to observe that the survival models of the subgroups that the EsmamDS delivers not only are more distinct from each other, but also capture a wider range of survival responses (from lower to higher survival curves).

Comparing our Diverse Search approach to its predecessor Esmam, we observe that the description language that constrain attributes on a set of values (instead of a single value) along with a subgroup selection method that prioritises generalisation yield more informative subgroups.
That is because the EsmamDS yields subgroups with statistically larger coverage but with similar description $length$.
Also, the heuristic function that improves search exploration and the minimisation of redundancy in the final subgroups yield a lower number of patterns that better represent the data (an increase of over 20\% of the data set coverage).

We observe high levels of redundancy in beam-search approaches, which is an inherent problem of this search heuristic.
By prioritising a number of best solutions in each search level, this heuristic leans towards regions of the search space that usually comprise a large number of variations of the same finding: many (slightly) different subgroup descriptions implying many (almost) equal subgroups that have (almost) equal target distribution \cite{SD:algorithm_DiverseSD}.
This problem can be observed in the high levels of descriptive and coverage redundancy ($\rho_D, \rho_C$).
Another indicative of such redundancy is descriptions with larger $length$ and smaller coverages, in most cases comprising refinements of a more general pattern. 
The high number of refinements among the final findings also reflects on low data representation, which is less than 50\% of the data set cases for this family of algorithms.
Additionally, such redundant patterns yield poor diversity of (singular) unusual survival behaviours ($\rho_M$) with over 50\% of the discovered survival models being similar to each other.

\begin{figure}[t]
    \centering
    \includegraphics[width=0.49\textwidth]{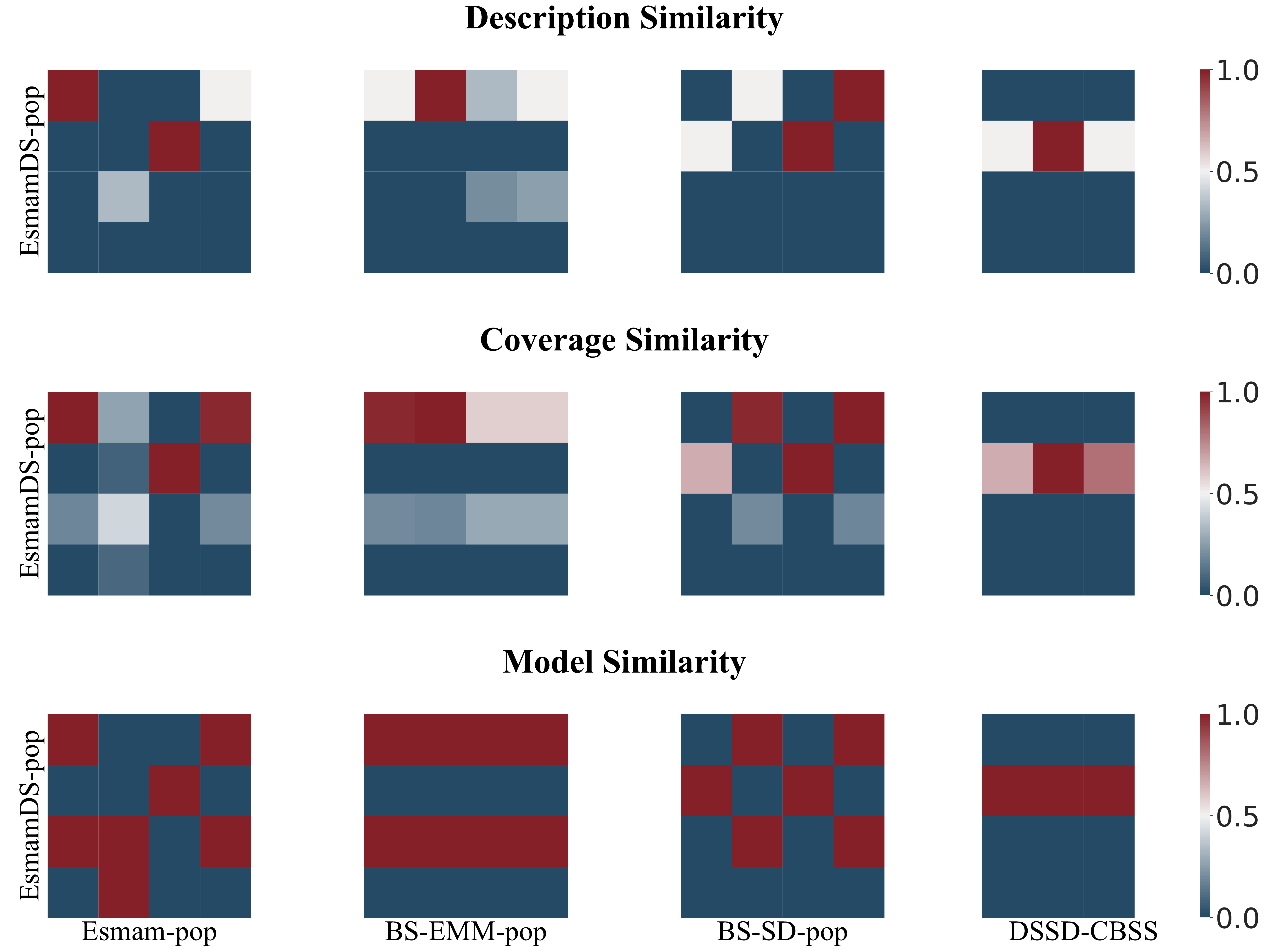}
    \caption{Similarity measures the EsmamDS subgroup set and the $\mathcal{B}$-\textit{population} compared approaches on \textit{mgus} dataset (\textit{experiment 0}): description similarity $sim_D$; coverage similarity $sim_C$; and model similarity $sim_M$.}
    \label{fig:heatmap_pop_btw}
\end{figure}

Finally, we use the similarity measures $sim_D$, $sim_C$ and $sim_M$ to assess the similarity between the set of subgroups delivered by the EsmamDS and those discovered by each compared approach.
For that matter, we compute the measures for all combinations of subgroups in both sets, and we present such analysis in a heatmap matrix.
\autoref{fig:heatmap_pop_btw} presents the three types of similarity for the $\mathcal{B}$-\emph{population} algorithms over \emph{mgus} data set (the whole set of comparisons may be viewed in our supporting \href{https://github.com/jbmattos/EsmamDS/tree/icde2022/experiments}{website}).
In each heatmap matrix, the rows represent the subgroups discovered by the EsmamDS algorithm and the columns are the unique subgroups discovered by the compared approach.
The horizontal comparisons in the figure show similarities between a single EsmamDS subgroup and all subgroups found by the other approaches. 
Most experiments present such patterns, indicating that the sets of subgroups delivered by EsmamDS somehow encompass the subgroups returned by the other algorithms.

To better analyse such results, we use the comparison between EsmamDS-pop and BS-EMM-pop (the second column of heatmaps in \autoref{fig:heatmap_pop_btw}) as an example.
From the plot, we can observe that all subgroups in the BS-EMM-pop set (columns) are somehow similar to the first EsmamDS subgroup (first row). 
Such subgroups' descriptions are as follows, where $G_0$ is the EsmamDS subgroup and $G^*$ are the BS-EMM-pop subgroups.
\begin{table}[H]
\begin{center}
\begin{tabular}{p{1.4cm} p{1.8cm} p{0.05cm} p{2cm} p{0.05cm} p{1.3cm}}
$G_{0}:$      & $age=[74,87]$ &     &                   &       &               \\
$G^*_{1}:$    & $age=[74,87]$ &     &                   &       &               \\
$G^*_{3}:$    & $age=[74,87]$ & $\land$ & $creat=0$         &       &               \\
$G^*_{0}:$    & $age=[74,87]$ & $\land$ & $pcdx=$\textit{ not-prog}   &       &               \\
$G^*_{2}:$    & $age=[74,87]$ & $\land$ & $pcdx=$\textit{ not-prog}   & $\land$   & $creat=0$\\
\end{tabular}
\end{center}
\end{table}

We can observe that all subgroups discovered by BS-EMM-pop are refinements of (or equal to) a single EsmamDS-pop subgroup.
Note that although the \emph{description similarity} measure captures intersections in subgroups' descriptions, such measure decreases as a refinement becomes deeper (e.g. $G^*_{1}, G^*_{3}$ and $G^*_{2}$, respectively), hence, low levels may reflect specialisations of a single pattern.
We also observe that the refinements of a subgroup yield only slight variations in \emph{cover similarity}.
Besides, when analysing the \emph{model similarity}, we can observe that all refinements delivered by the BS-EMM-pop algorithm present survival models similar to their generalisation, $G_{0}$.
That is because subsets of a subgroup's coverage (or refinements of a subgroup's description) may present exceptional behaviour when compared to the baseline model but not necessarily yield a distinct distribution from their generalisation.

There are cases where we observe that different descriptions yield some degree of cover and model similarities.
However, most of the times, such horizontal patterns in the heatmaps indicate the presence of subgroup refinements that can be generalised by a single (or few) subgroup(s) in the EsmamDS set without loss of information.
Hence, when comparing the set of subgroups found by each approach, the EsmamDS usually represents the majority of the findings delivered by the other algorithms in a more compact and general way, while providing higher diversity of the discovered patterns and models.
%
%
%
\section{Conclusions}
\label{sec:conclusions}
This paper presents the EsmamDS, an EMM framework based on Ant-Colony Optimisation (ACO) meta-heuristics for mining subgroups presenting unusual survival behaviour.
The EsmamDS builds on the Esmam algorithm to tackle the problem of pattern redundancy.
The approach presented uses the ACO search mechanisms to improve search exploration and a new subgroup selection method to provide a set of subgroups that minimises redundancy in description, coverage and model. 
Additionally, EsmamDS descriptions allows attributes to be constrained on a set of values, improving the generalisation and comprehensibility of the discovered patterns. 

The EsmamDS was confronted with other approaches to mine subgroups with disparities in survivability on 14 survival data sets. 
The experiments showed that the EsmamDS yields smaller sets of subgroups, with simpler and more informative characterisations, capable of representing the majority of the data observations.
The set of subgroups discovered by our approach usually encompasses the subgroups delivered by the others, but with more general and compact representations, while discovering subgroups that the others do not uncover.
The EsmamDS also minimises the presence of subgroup refinements by allowing their occurrence only if they present distinct survival distribution, providing subgroups that are more diverse concerning their description and coverage while delivering a variety of interesting survival models.

The main limitations of EsmamDS are not to assure the statistical robustness of the findings and not to work with numerical attributes.
Thus, adaptations to handle large and high-dimensional data are essential to investigate a broader scope of problems.
Additionally, the ACO search mechanisms may be adapted to enrich the subgroup mining procedure to consider correlations between features and to incorporate more complex data which representation cannot be considered in the domain of descriptive attributes without loss of comprehensibility.
\section*{Acknowledgment}
This study was financed in part by the Coordenação de Aperfeiçoamento de Pessoal de Nível Superior - Brasil (CAPES) - Finance Code 001, by the National Council for Scientific and Technological Development -- CNPq, and by the Google Latin America Research Awards 2020.
%
\bibliographystyle{IEEEtranN}    
\bibliography{references}       

\end{document}